\pdfoutput=1
\documentclass[11pt]{article}

\usepackage[final]{acl}

\usepackage{times}
\usepackage{latexsym}
\usepackage{adjustbox}
\usepackage{float}
\usepackage{placeins}

\usepackage[T1]{fontenc}
\usepackage[utf8]{inputenc}

\usepackage{microtype}

\usepackage{inconsolata}

\usepackage{graphicx}
\usepackage[table]{xcolor}
\usepackage{multirow} 
\usepackage{enumitem}
\setitemize{noitemsep,topsep=0pt,parsep=0pt,partopsep=0pt}
\usepackage{url}

\usepackage{dialogue}
\usepackage{amsthm}
\theoremstyle{definition}
\newtheorem{example}{Example}

\title{"Mm, Wat?" Detecting Other-initiated Repair Requests in Dialogue}

\author{
 \textbf{Anh Ngo\textsuperscript{1,5}},
 \textbf{Nicolas Rollet\textsuperscript{1,2}},
 \textbf{Catherine Pelachaud\textsuperscript{4}},
 \textbf{Chloé Clavel\textsuperscript{1,3}},
\\
\\
 \textsuperscript{1}ALMAnaCH, INRIA Paris ,
 \textsuperscript{2}Télécom Paris, SES, Institut Polytechnique de Paris, I3-CNRS,
 \\
 \textsuperscript{3}Télécom Paris, LTCI, Institut Polytechnique de Paris,
 \textsuperscript{4}CNRS, ISIR, Sorbonne University,
 \\
 \textsuperscript{5}ISIR, Sorbonne University
\\
 \small{
    \url{{anh.ngo-ha, nicolas.rollet, chloe.clavel}@inria.fr, catherine.pelachaud@upmc.fr}
 }
}

\begin{document}
\maketitle
\begin{abstract}
%  to be filled
Maintaining mutual understanding is a key component in human-human conversation to avoid conversation breakdowns, in which repair, particularly Other-Initiated Repair (OIR, when one speaker signals trouble and prompts the other to resolve), plays a vital role. However, Conversational Agents (CAs) still fail to recognize user repair initiation, leading to breakdowns or disengagement. This work proposes a multimodal model to automatically detect repair initiation in Dutch dialogues by integrating linguistic and prosodic features grounded in Conversation Analysis. The results show that prosodic cues complement linguistic features and significantly improve the results of pretrained text and audio embeddings, offering insights into how different features interact. Future directions include incorporating visual cues, exploring multilingual and cross-context corpora to assess the robustness and generalizability.
\end{abstract}

\section{Introduction}
\label{introduction}
\begin{figure}[h]
  \centering
  \includegraphics[width=\columnwidth]{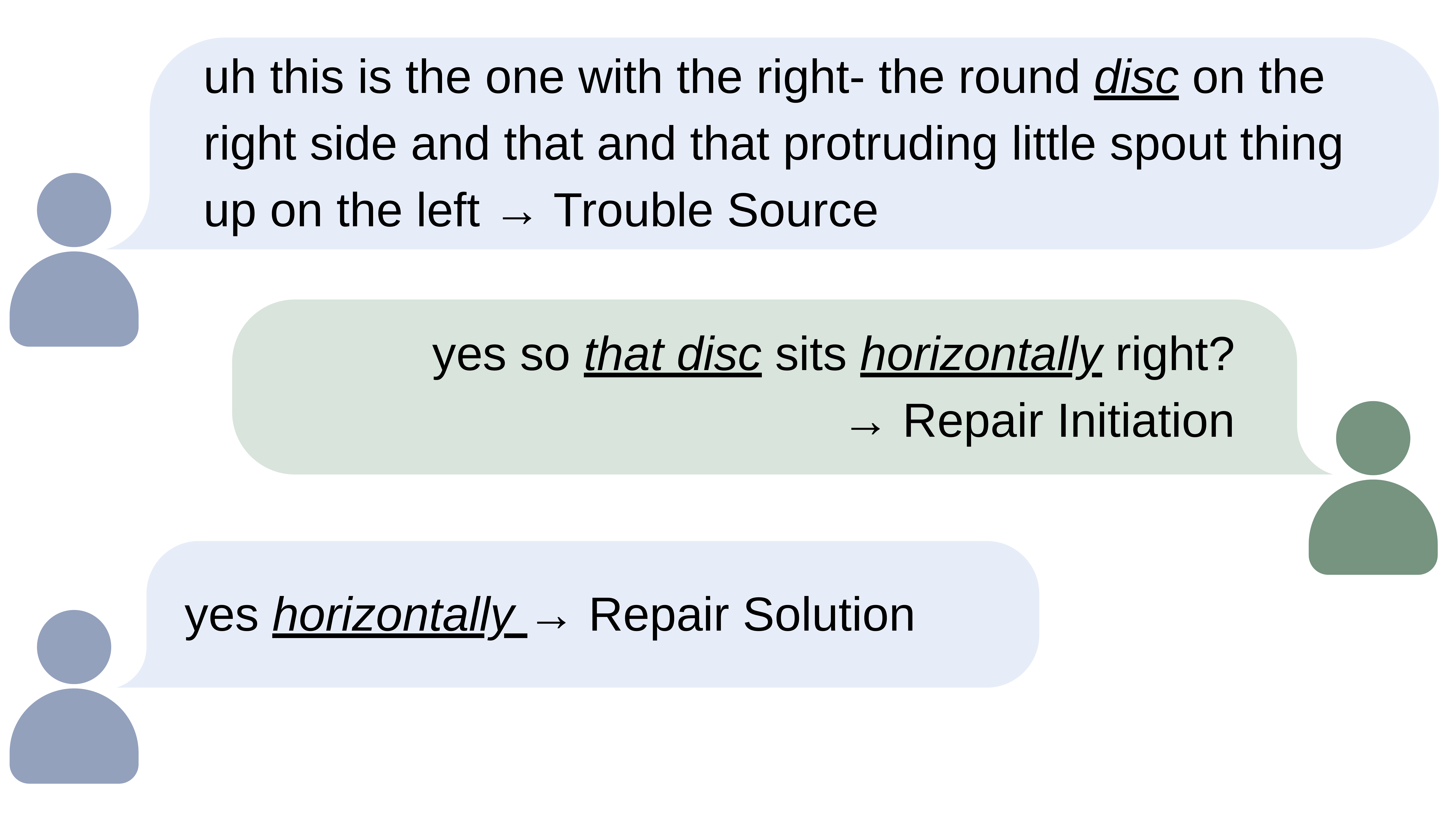}
  \caption{Other-initiated Repair (OIR) sequence example from \citet{Rasenberg2022}, English translated: repair initiation (green) signals trouble of ambiguous object reference \emph{disc} with candidate understanding \emph{horizontally}, confirmed by repair solution \emph{yes horizontally}.}
  \label{fig:intro_example}
\end{figure}

Conversational agents (CAs), software systems that interact with users using natural language in written or spoken form, are increasingly being used in multiple domains such as commerce, healthcare, and education \cite{Allouch2021}. While maintaining smooth communication is crucial in these settings, current state-of-the-art (SOTA) CAs still struggle to handle conversational breakdowns. Unlike humans, who rely on conversational repair to resolve issues like mishearing or misunderstanding \cite{Schegloff1977, Schegloff2000}, CAs' repair capabilities remain limited and incomplete. Repair refers to the interactional effort by which participants suspend the ongoing talk to address potential trouble, which can be categorized by who initiates and who resolves it: the speaker of the trouble (self) or the co-participant (other) \citep{Schegloff2000}. This work focuses on \textbf{Other-initiated Self-repair}, in short, \textbf{Other-initiated Repair (OIR)}, where the \textbf{talk of a speaker is treated as problematic by a co-participant via repair initiation, and the original speaker resolves it, as illustrated in Figure~\ref{fig:intro_example}.} Current CAs handle repairs in a limited fashion that mainly support self-initiated repair by the agent (e.g., the agent asks users to repeat what they said) \cite{Li2020, Cuadra2021, Ashktorab2019} or rely on user self-correction when users realize troubles and clarify their own intent (e.g., saying "no, I mean…") \cite{balaraman2023}. However, CAs struggle to recognize when users signal trouble with the agent's utterances (other-initiated) and fail to provide appropriate repair (self-repaired), while effective communication requires bidirectional repair capabilities \cite{Moore2024}. Supporting this, \citet{Gehle2014} found that robots failing to resolve communication issues quickly caused user disengagement, while \citet{arkel2020} showed that basic OIR mechanisms improve communication success and reduce computational and interaction costs compared to relying on pragmatic reasoning.

Modeling OIR strategies on CAs that recognize user‑initiated repair first requires robust automatic repair initiation detection in human-human interaction. Previous work has established foundations for text‑based approaches, training with English corpora, and relying on lexical cues \cite{Hohn2017, Purver2018, Alloatti2024}. However, prosodic cues tend to be more cross-linguistically stable than surface forms \cite{Dingemanse2015, Trevor2013, Walker2017}, and can provide valuable insight into the pragmatic functions of expressions like the interjection "huh". Building upon text-based methods, this work focuses on spoken dialogue interaction, where prosodic cues provide additional signals for repair initiation detection that may be missed by text-only models trained on transcriptions. Finally, understanding the OIR sequence also requires examining the local sequential environment of the surrounding turns, which we term "dialogue micro context", based on \citet{Schegloff1987}'s work on local interactional organization.

%while “what?” varies from English wat to Siwu be: and Yélî Dnye lukwe, the interjection huh? retains a similar intonation across languages \cite{Dingemanse2015, Trevor2013, Walker2017}, emphasizing the limitation of relying only on textual patterns. Finally, understanding OIR sequence also requires examining the local sequential environment of surrounding turns, which we term "dialogue micro context" \cite{Schegloff2000}.

These gaps motivate our main research question: \textbf{What are the verbal and prosodic indicators of repair initiation in OIR sequences and how can we model them?} To address this, we analyze OIR sequences in a Dutch task-oriented corpus, focusing on text and audio patterns where one speaker initiates repair. Drawing on Conversation Analysis literature, we introduce feature sets and a computational model to detect such requests. Our contributions are in two folds: (1) a novel multimodal model for detecting repair initiations in OIR sequences that integrates linguistic and prosodic features extracted automatically based on the literature, advancing beyond text- or audio-only approaches; (2) provide insights into how linguistic and prosodic features interact and contribute in detection performance, grounded in Conversation Analysis, and what causes model misclassifications. The remainings of this paper is structured as follows: Section~\ref{related_works} reviews SOTA computational models for OIR detection and related dialogue understanding tasks. Section~\ref{background} provides the used OIR coding schema and typology, and Section~\ref{approach} details our approach, including linguistic and prosodic feature design. Section~\ref{experiments_results} presents our experiment details and results, followed by error analysis in Section~\ref{error_analysis}.

\section{Related Work}
\label{related_works}
An early approach to automatic OIR detection was proposed by \citet{Hohn2017}, with a pattern-based chatbot handling user-initiated repair in text chats between native and non-native German speakers. \citet{Purver2018} extended this by training a supervised classifier using turn-level features in English, including lexical, syntactic, and semantic parallelism between turns. More recently, \citet{Alloatti2024} introduced a hierarchical tag-based system for annotating repair strategies in Italian task-oriented dialogue, distinguishing between utterance-specific and context-dependent functions. Closely related, \citet{GariSoler2025}'s recent work introduced and investigated the task of automatically detecting word meaning negotiation indicators, where speakers signal a need to clarify or challenge word meanings, a phenomenon that can be seen as a specific form of repair initiation.
%An early attempt in automatic OIR detection was introduced by \citet{Hohn2017} with a pattern-based chatbot capable of handling user-initiated repair in text-based chat between native and non-native German speakers. Building on this, \citet{Purver2018} defined turn-level features from conversation transcripts to train a supervised classifier, including lexical, syntactic, and semantic parallelism between turns. More recently, \citet{Alloatti2024} proposed a hierarchical tag-based system to annotate repair strategies in task-oriented conversations, distinguishing between utterance-specific (Inherent) and context-dependent (Backward) functions

Although direct research on OIR detection is still limited, advances in related dialogue understanding tasks provide promising foundations for our work. \citet{Miah2024} combined pretrained audio (Wav2Vec2) and text (RoBERTa) embeddings to detect dialogue breakdowns in healthcare calls. Similarly, \citet{Huang2023} used BERT, Wav2Vec2.0, and Faster R-CNN for intent classification, introducing multimodal fusion with attention-based gating to balance modality contributions and reduce noise. \citet{Saha2020b} proposed a multimodal, multitask network jointly modeling dialogue acts and emotions using attention mechanisms. More recently, high-performing but more opaque and resource-intensive approaches have emerged, such as \citet{Mohapatra2024} showed that larger LLMs outperform smaller ones on tasks like repair and anaphora resolution, albeit with higher computational cost and latency.

Despite robust performance, recent largest models remain difficult to interpret due to their black-box nature and multimodal fusion complexity \cite{Jain2024}. To address this gap, we propose a computational model for repair initiation detection in Dutch spoken dialogue that fuses pretrained text and audio embeddings with linguistic and prosodic features \textbf{grounded in Conversation Analysis}. The model also integrates a multihead attention mechanism to weigh and capture nonlinear relationships across modalities, allowing our model to keep the strengths of multimodal deep learning while \textbf{offering insight from linguistic and prosodic features to understand their interaction and impact} towards model's decision.
%To address these gaps, we propose a \textbf{computational model for OIR request detection that fuses pretrained text and audio embeddings with linguistic and prosodic features grounded in Conversation Analysis}, allowing our model to keep the strengths of multimodal deep learning while \textbf{offering insight from linguistic and prosodic features to intepret their interaction and impact} towards model's decision.
\begin{figure*}[h]
  \centering
  \includegraphics[width=0.9\textwidth]{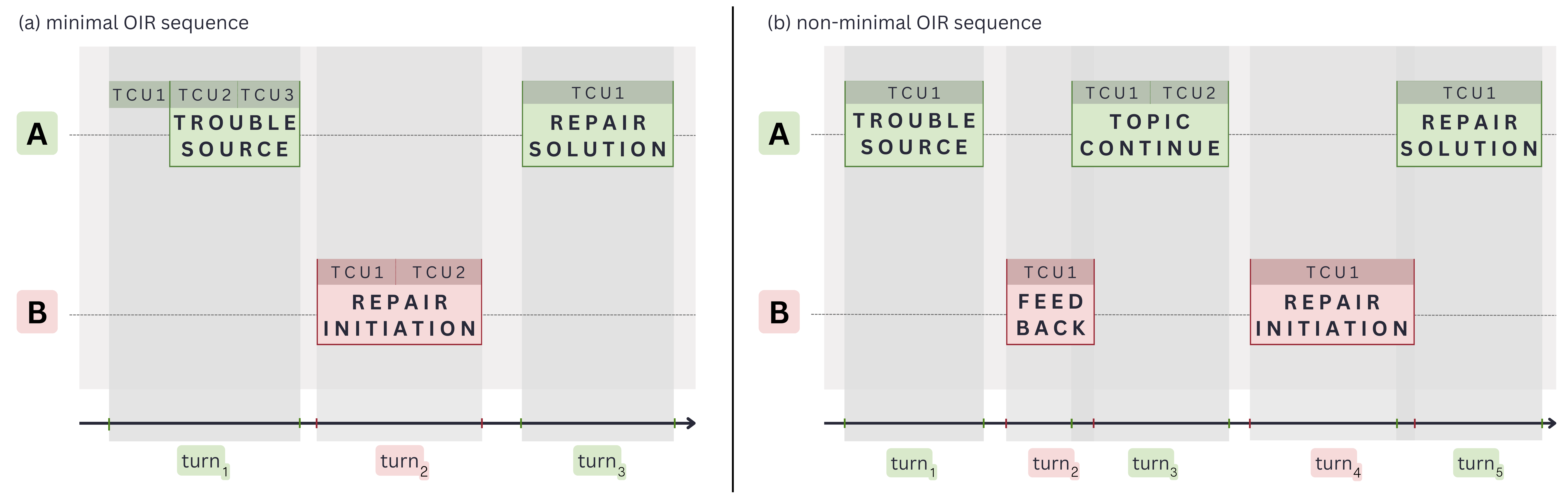}
  \caption{OIR sequence organization between 2 speakers A (green) and B (red): (a) Minimal; (b) Non-minimal}
  \label{fig:OIR_sequences}
\end{figure*}

\section{OIR Coding Schema and Typology}
\label{background}

We follow \citet{Dingemanse2015}'s coding schema, which structures OIR sequences into three components: trouble source, repair initiation, and repair solution segments, with repair initiation categorized as \emph{open request} (the least specific, not giving clues of trouble), \emph{restricted request} (implied trouble source location), or \emph{restricted offer} (the most specific, proposing a candidate understanding). Throughout this work, repair initiation refers specifically to this component within OIR sequences. We use the corpus and the OIR sequences annotation from \citet{Rasenberg2022}, where dialogues were manually transcribed and segmented into Turn Construction Units (TCUs), the smallest meaningful elements of speech that can potentially complete a speaker turn. They align OIR component boundaries with these pre-existing TCU boundaries. Following the conversational analysis practice, such as in \cite{Mondada2018}, we adopt the “segment” as our unit of analysis, defined as: stretches of talk corresponding to annotated OIR components (e.g., repair initiation) that may span one or more TCUs within larger speaker turns (illustrated in Figure~\ref{fig:OIR_sequences}). This allows us to target only the stretch of talk relevant to the OIR component, avoiding the overinclusiveness of full turns. Figure~\ref {fig:OIR_sequences} illustrates two organizational scenarios of OIR sequences described in \citet{Dingemanse2015}, including: \textit{minimal} (repair initiation produced immediately after the turn containing the trouble source) and \textit{non-minimal} (repair initiation delayed by a few turns).

\section{Proposed Approach}
\label{approach}
%\subsection{Task Formulation \& Overview Architecture}
\subsection{Overview}
\begin{figure*}[ht]
  \centering
  \includegraphics[width=0.9\textwidth]{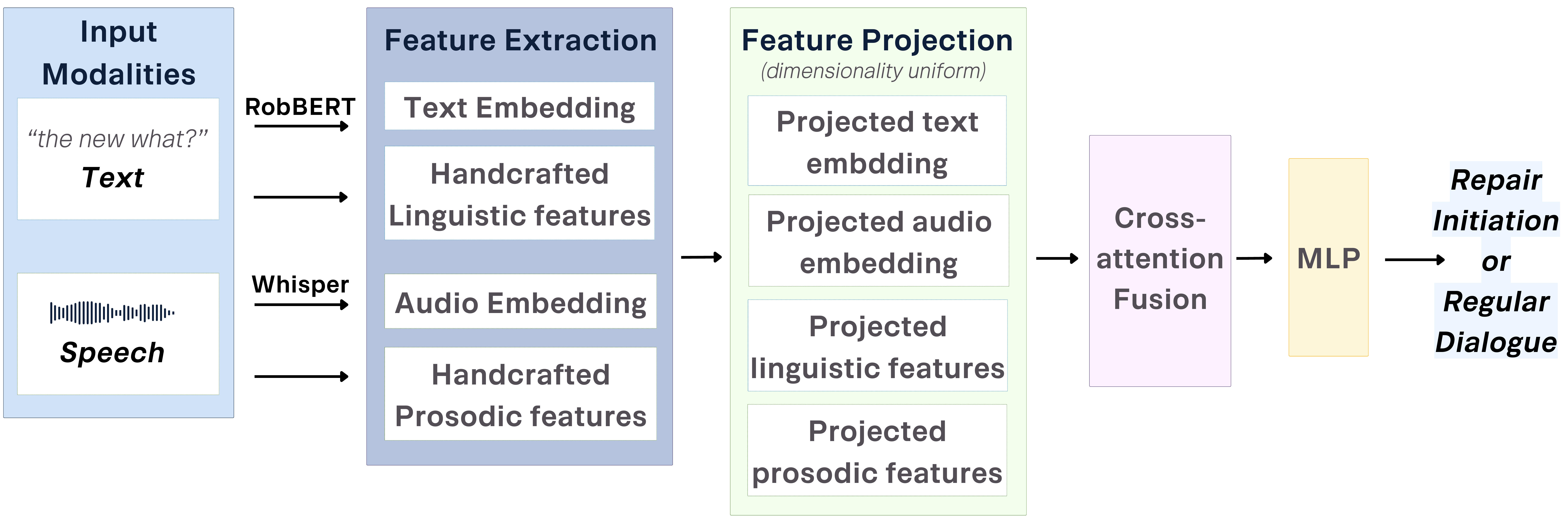}
  \caption{Multimodal architecture for repair initiation detection}
  \label{fig:model_overview}
\end{figure*}
\paragraph{Task Formulation.} We formulate the repair initiation detection task as a binary classification problem. Given a segment (\(x_i\)), corresponding to one or several TCUs within a speaker turn, the task is to predict whether it is an OIR repair initiation or a regular dialogue (RD) segment (\textit{i.e.}, not belonging to an OIR sequence). In this initial study, we limit the scope to detecting repair initiations only, without classifying other OIR components such as trouble sources or repair solutions. This simplification allows us to establish a baseline for the most critical component in the OIR sequence, the moment when repair is initiated by another speaker.
\paragraph{Architecture Overview.} Figure~\ref{fig:model_overview} shows our proposed multimodal approach for repair initiation detection. We incorporate the handcrafted linguistic and prosodic features, automatically computed based on literature reviews, with embeddings from pretrained models (RobBERT for text, Whisper for audio). For a given segment (\(x_i\)), we  first extract both handcrafted features and pretrained embeddings from text and audio modalities. All features are then projected to a shared dimensionality to ensure consistency across modalities. To capture the complex interactions between text and audio embeddings with handcrafted features, a multihead attention mechanism was employed to weigh and capture nonlinear relationships. Finally, the whole representation is obtained by concatenating the text embedding and the fused representation from multihead attention.

\subsection{Pretrained Models}
\paragraph{Language model.} Our proposed approach utilizes BERT \cite{Devlin2019}, a transformer-based language model, to obtain text embedding of the current given segment. As our corpus is in Dutch, we use the pretrained RobBERT \cite{Delobelle2020} model, which is based on the BERT architecture, pretrained with a Dutch tokenizer, and 39 GB of training data. We use the latest release of \emph{RobBERT-v2-base} model which pretrained on the Dutch corpus OSCAR 2023 version, which outperforms other BERT-based language models for several different Dutch language tasks.

\paragraph{Audio model.} For audio representations, we utilize Whisper \cite{Radford2023}, an encoder-decoder transformer-based model trained on 680,000 hours of multilingual and multitask speech data, to extract audio embeddings from our dialogue segments. Whisper model stands out for its robustness in handling diverse and complex linguistic structures, a feature that is crucial when dealing with Dutch, a language known for its intricate syntax. Besides, Whisper was trained on large datasets including Dutch and demonstrated good performance in zero shot learning, making it ideal serving as a naive baseline for task with small corpus like ours.

\subsection{Dialogue Micro Context}
% context integration approach
\citet{Schegloff1987} demonstrated that the OIR sequence is systematically associated with multiple organizational aspects of conversation, and understanding an OIR repair initiation requires examining the local sequential environment, which he terms the micro context, that we adopt in this work. Therefore, for each given target segment \(x_{i}\), to capture the micro context, we iteratively concatenate the previous (\(x_{i-j}\)) and following (\(x_{i+j}\)) segment within a window of size (\(j\)), using special separator token of transformers (e.g. [SEP] for BERT-based models) until reaching the maximum token limit (excluding [CLS] and [EOS]), inspired by similar ideas in \cite{Wu2020}. If the sequence exceeds the limit, we truncate the most recently added segments. The final sequence is enclosed with [CLS] and [EOS], as shown in Figure~\ref{fig:dialogue_context_concatenation} (Appendix~\ref{sec: micro_context}). 

\subsection{Linguistic Feature Extraction}

\begin{figure*}[ht]
  \centering
  \includegraphics[width=\textwidth]{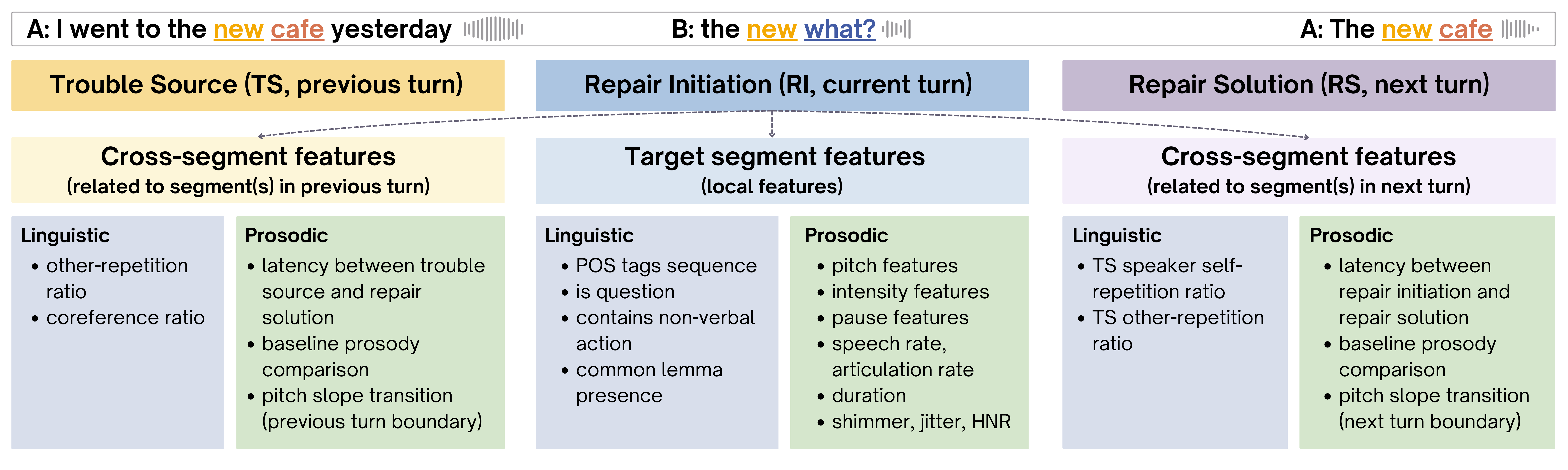}
  \caption{Handcrafted linguistic and prosodic features design}
  \label{fig:feat_design}
\end{figure*}

% handcrafted linguistic
Figure~\ref{fig:feat_design}(a) outlines our linguistic feature set for the representation of the target segment, capturing local properties such as part-of-speech (POS) tagging patterns, question formats, transcribed non-verbal actions (target segment features), and features, which quantify repetition and coreference across turns to reflect backward and forward relations around the repair initiation (cross-segment feature to capture micro context). The detailed description is in the Appendix~\ref{sec:appendix_linguistic}. 

% \begin{figure}[h]
%   \centering 
%   \includegraphics[width=\columnwidth]{feat_linguistic_design.pdf}
%   \caption{Handcrafted linguistic features set}
%   \label{fig:feat_linguistic_design}
% \end{figure}

\subsubsection{Target Segment Features}
We automatically extracted the linguistic features proposed by \cite{Ngo2024} at the intra-segment level to capture grammatical and pragmatic patterns related to the repair initiation. For instance, \cite{Ngo2024} shows that restricted requests often show a POS tag sequence pattern of interrogative pronouns followed by verbs, while OIR open requests and regular dialogue segments differ in key lemmas used of the same tag: modal auxiliary verb kunnen (“can”) vs. primary auxiliary verb zijn (“to be”). We also include question mark usage, derived from original transcription, which is marked with a question mark if the annotator detected question prosody. It implicitly reflects prosodic cues as interpreted by the human annotators, which are relevant to repair initiation, as described in \citet{Schegloff2000} regarding interrogative and polar question formats. A complete list of features is fully provided in Appendix~\ref{sec:appendix_linguistic}.

\subsubsection{Cross-Segment Features} 
Grounded on the literature \cite{Schegloff2000, Ngo2024}, we define inter-segment features that capture the sequential dynamics of the repair initiation, including repetitions and the use of coreferences referring to entities in prior turns containing the trouble source segment. We also compute self and other-repetition in the subsequent turn containing the repair solution segment, to capture how the trouble source speaker responds. These features reflect the global dynamics of OIR sequences.

\subsection{Prosodic Features Extraction}
% handcrafted prosodic
Prosody plays a crucial role in signaling repair initiation. Previous studies in Conversation Analysis show that pitch, loudness, and contour shape can indicate whether repair initiation is perceived as "normal" or expresses "astonishment"\cite{Selting1996}, and that Dutch question types differ in pitch height, final rises, and F0 register \cite{Haan1997}. Building upon these characteristics, we design a prosodic feature set that includes both local features within the target segment, such as pitch, intensity, pauses, duration, and word-level prosody, and global features across segments of the OIR sequence, such as latency between OIR sequence segments, pitch slope transitions at boundaries, and comparison to speaker-specific prosodic baselines. The features are detailed in Figure~\ref{fig:feat_design}(b) and in the Appendix~\ref{sec:appendix_prosodic}.

% \begin{figure}[h]
%   \centering
%   \includegraphics[width=\columnwidth]{feat_prosodic_design.pdf}
%   \caption{Handcrafted prosodic feature set}
%   \label{fig:feat_prosodic_design}
% \end{figure}

\subsubsection{Target Segment Features}
We use Praat \cite{Boersma2000} to extract prosodic features at the segment level, including: pitch features (e.g., min, max, mean, standard deviation, range, number of peaks) which are computed from voiced frames after smoothing and outlier removal, with pitch floor/ceiling set between 60–500 Hz and adapted to each speaker range \cite{Bezooijen1995, Theelen2017, Connell2024}; first (mean and variability of pitch slope change) and second derivatives (pitch acceleration) of pitch contour, capturing pitch dynamics. Additional features are intensity (e.g., min, max, mean, range, standard deviation), and voice quality measures (jitter, shimmer, and harmonics-to-noise ratio). We also model pause-related features by detecting silent pauses over 200 ms and categorizing them by duration and position in the utterance, reflecting their conversational function associated with repair possibilities \cite{Donzel1996, Elliott2018}. Inspired by findings about prosody of other-repetition in OIR sequences \cite{Dingemanse2015b, Walker2017}, we extract pitch and intensity features for repeated words from the trouble source segment, and for the specific repair marker "wat" (what/which/any), as indicators of repair initiation type and speaker perspective \cite{Huhtamaki2015}.

\subsubsection{Cross-Segment Features} 
To model the speaker-specific prosodic variation \cite{Bezooijen1995, Theelen2017, Connell2024}, we normalize pitch and intensity using z-scores, relative percentage change, and position within the speakers' range. These features capture how far the current segment deviates from the speaker's typical behaviour across previous turns and the normalized range position of the current segment within the speaker's baseline. Inspired by work on prosodic entrainment \cite{Levitan2011}, we also compute pitch and intensity slope transitions across segment boundaries (e.g., TS→OIR, OIR→RS), both within and across speakers, to assess prosodic alignment. We normalized slopes to semitones per second for consistency across speakers.

\section{Experiments \& Results}

\begin{table*}[!ht]
\centering
    \resizebox{0.8\textwidth}{!}{%
    \begin{tabular}{lcccc}
    \hline
    \textbf{Model} & \textbf{Modal \& Features} & \textbf{Precision} & \textbf{Recall} & \textbf{F1-score} \\
    \hline
    Text\textsubscript{Emb}         & U \& T       & $72.0 \pm 4.0$ & $87.6 \pm 7.5$ & $78.9 \pm 4.7$ \\
    Audio\textsubscript{Emb}        & U \& A       & $72.6 \pm 9.7$  & $76.3 \pm 13.1$ & $70.6 \pm 8.1$  \\
    \rowcolor{gray!15}
    Multi\textsubscript{Emb}        & M \& T+A     & $79.1 \pm 5.4$  & $82.2 \pm 3.8$ & $82.1 \pm 0.9$  \\
    \hline
    Text\textsubscript{Ling}        & U \& L       & $82.2 \pm 3.6$  & $80.4 \pm 6.1$ & $80.4 \pm 3.8$  \\
    Audio\textsubscript{Pros}       & U \& P       & $81.7 \pm 4.2$  & $77.4 \pm 5.4$ & $77.3\pm2.7$ \\
    \rowcolor{gray!15}
    Multi\textsubscript{LingPros}   & M \& L+P     & $81.7\pm7.6$  & $82.2\pm1.5$ & $81.8\pm3.4$  \\
    \hline
    \rowcolor{gray!30}
    Multi\textsubscript{Ours}       & M \& T+A+L+P & $\mathbf{93.2\pm2.8}$ & $\mathbf{96.1\pm2.6}$ & $\mathbf{94.6\pm2.3}$ \\
    \hline
    \end{tabular}%
    }    \captionsetup{justification=raggedright,singlelinecheck=false}
    \caption*{\small \textbf{U}: Unimodal, \textbf{M}: Multimodal, \textbf{T}: Text, \textbf{A}: Audio, \textbf{P}: Prosodic features, \textbf{L}: Linguistic features}
      \caption{Overall results across modalities for repair initiation detection. The table groups models by research question: \textbf{RQ1} compares unimodal vs.\ multimodal combinations of audio and text; \textbf{RQ2} compares handcrafted features with pretrained embeddings.}
      \label{tab:overall_results}
\end{table*}

\label{experiments_results}
To answer the main research question mentioned in Section~\ref{introduction}, we design the experiments to answer the following research sub-questions: \textit{i)} \textbf{RQ1}: To what extent do audio-based features complement text-based features in identifying repair initiation? \textit{ii) }\textbf{RQ2}: Do our proposed linguistic and prosodic features (see Figures~\ref{fig:feat_design}(a) and \ref{fig:feat_design}(b)) perform better than pretrained embeddings? \textit{iii) }\textbf{RQ3}: Which prosodic and linguistic features contribute the most to repair initiation detection? \textit{iv) }\textbf{RQ4}: How does the involvement of dialogue micro context affect detection performance?

\subsection{Implementation Details}
  \paragraph{Dataset.} Based on \cite{Colman2011}'s finding that repair occurs more frequently in task-oriented dialogues, we selected a Dutch multimodal task-oriented corpus \cite{Rasenberg2022, Eijk2022}, containing 19 dyads collaborating on referential communication tasks in a standing face-to-face setting. For each round, participants alternated roles to describe (Director) or identify (Matcher) a geometric object (called "Fribbles") displayed on screens, in which the unconstrained design encouraged natural modality use and OIR sequences. \citet{Rasenberg2022} annotated OIR sequences using \citealp{Dingemanse2015}'s schema, resulting in 10 open requests, 31 restricted requests, and 252 restricted offers. While we acknowledge that OIR sequences are rarer in natural dialogue, our goal in this paper is to study detection performance with sufficient examples of both classes. Therefore, we balanced the dataset with 306 randomly selected regular dialogue segments, stratified across all dyads, resulting in 712 samples overall. The data were split 70:15:15 for training, validation, and testing. Limitations regarding the generalizability of the artificial balancing are discussed in Section~\ref{sec:limitations}. Examples of Fribbles objects and repair initiation types are provided in the Appendix~\ref{sec:corpus_info} and \ref{sec:oir_types_examples}.
  \paragraph{Training Details.} We fine-tuned our models using 10-fold cross-validation, in which the optimal learning rate was 2e-5. We employed AdamW optimizer with weight decay of 0.01 and a learning rate scheduler with 10\% warm-up steps. Training ran for up to 20 epochs with 3-epoch early stopping patience, and batch size 16. The source code is publicly available \footnote{\url{https://github.com/haanh764/other_initiated_repair_detection}}.
  \paragraph{Evaluation Metrics.} We evaluated model performance using binary classification metrics including precision, recall, and macro F1-score.

\subsection{Experiment Scenarios \& Results Analysis}

%\subsubsection{RQ1: Audio vs.\ Text Complementarity} 
\paragraph{RQ1: Audio vs.\ Text Complementarity.}
To address RQ1, we compare the performance of unimodal against multimodal models, including: \textit{i)} Single \textbf{Text\textsubscript{Emb}} or \textbf{Audio\textsubscript{Emb}} vs.\ \textbf{Multi\textsubscript{Emb}}; \textit{ii)} Single \textbf{Text\textsubscript{Ling}} or \textbf{Audio\textsubscript{Pros}} \textit{vs.}\ \textbf{Multi\textsubscript{LingPros}}.
We examine whether integrating the audio-based features, either by pretrained embeddings or by using handcrafted prosodic features, will improve the performance of the text-based models. The multimodal models include \textbf{Multi\textsubscript{Emb}}, which fuses pretrained text and audio embeddings, and \textbf{Multi\textsubscript{LingPros}}, which combines handcrafted linguistic and prosodic features, using cross-attention fusion as illustrated in Figure~\ref{fig:model_overview}. 

From Table~\ref{tab:overall_results}, we observe that multimodal models consistently outperform unimodal ones across all metrics. For both pretrained embeddings and handcrafted features, text-based models outperform audio-based ones individually. However, incorporating audio improves performance in both settings. Specifically, in the pretrained setting, the multimodal model \textbf{Multi\textsubscript{Emb}} achieves an F1-score of 82.1, improving over \textbf{Text\textsubscript{Emb}} by 3.2 percentage points (pp) and over \textbf{Audio\textsubscript{Emb}} by 11.5 pp. Similarly, in the handcrafted feature setting, combining linguistic and prosodic features \textbf{Multi\textsubscript{LingPros}} yields an F1 of 81.8, outperforming Text\textsubscript{Ling} by 1.4 pp and \textbf{Audio\textsubscript{Pros}} by 4.5 pp. Interestingly, the unimodal handcrafted models \textbf{Text\textsubscript{Ling}}, \textbf{Audio\textsubscript{Pros}} show higher precision than recall, whereas \textbf{Multi\textsubscript{LingPros}} shows slightly higher recall, suggesting a tendency to favor detection over omission. This is potentially beneficial in interactive systems where missing an repair initiation could be more disruptive than a false alarm. For embedding-based models, recall exceeds precision in all cases, but the multimodal model shows a notable gain in precision, indicating a better trade-off between identifying true repair initiation and minimizing false positives.

%\subsubsection{RQ2: Handcrafted Features vs.\ Pretrained Embeddings}
\paragraph{RQ2: Handcrafted Features vs.\ Pretrained Embeddings.}
To address RQ2, we compare the performance of models using handcrafted features against the models using embeddings from pretrained models. We thus compare: \textit{i)} Text representations: text embeddings (\textbf{Text\textsubscript{Emb}}) \textit{vs.}\ handcrafted linguistic features (\textbf{Text\textsubscript{Ling}}); \textit{ii)} Audio representations: audio embeddings (\textbf{Audio\textsubscript{Emb}}) \textit{vs.}\ handcrafted prosodic features (\textbf{Audio\textsubscript{Pros}}); \textit{iii)} Combined approaches: multimodal models using pretrained embeddings (\textbf{Multi\textsubscript{Emb}}) \textit{vs.}\ using handcrafted linguistic and prosodic features (\textbf{Multi\textsubscript{LingPros}}) and \textit{vs.}\ our proposed approach leveraging both of them \textbf{Multi\textsubscript{Ours}}.

Table~\ref{tab:overall_results} shows that handcrafted feature models are comparable to embedding-based approaches. In unimodal settings, Text\textsubscript{Ling} achieves higher precision (+10 pp) with comparable F1-score (+1.5 pp) to \textbf{Text\textsubscript{Emb}}, despite lower recall (-7.2 pp). Likewise, \textbf{Audio\textsubscript{Pros}} outperforms \textbf{Audio\textsubscript{Emb} } across all metrics (precision +9.1 pp, recall +1.1 pp, F1-score +6.7 pp).  In multimodal settings, \textbf{Multi\textsubscript{Emb}} and \textbf{Multi\textsubscript{LingPros}} perform nearly identically (F1-score difference of 0.3 pp). Overall, we observe a general trend emerges: embedding-based approaches tend to achieve higher recall but lower precision, likely because they can learn more complex representation that captures more subtle patterns, whereas handcrafted feature models target specific repair initiation markers, such as question forms, repetition, and pause patterns, resulting in better balanced precision-recall trade-offs. The embedding models may also overgeneralize in the case of our small, task-specific corpus.

\begin{figure}[h]
  \centering
  \includegraphics[width=\columnwidth]{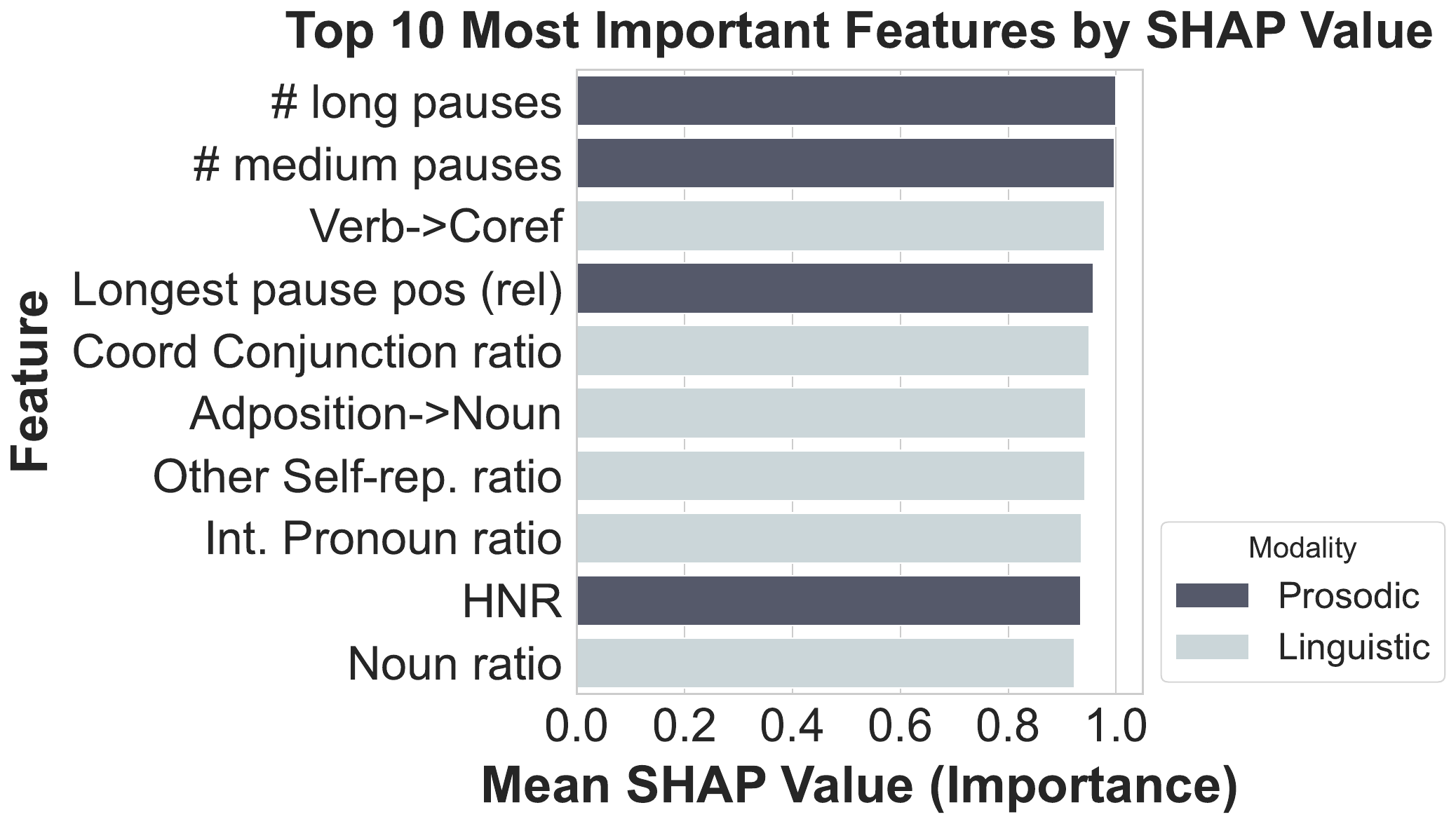}
  \caption{The top 10 most important handcrafted features ranked by SHAP value. Appendix~\ref{sec:appendix_feature_important} provides the full list of the 20 most contributed features.}
  \label{fig:feat_important_top10}
\end{figure}

\paragraph{RQ3: Handcrafted Feature Importance Analysis.}
Although the linguistic and prosodic features could not solely outperform pretrained text and audio embeddings, they are useful in interpreting the model's behaviours, especially to see if they are aligned with the Conversation Analysis findings. To answer RQ3, we used SHAP (SHapley Additive exPlanations) analysis to analyze the contribution and behaviours of linguistic and prosodic features towards the model's decision. Figure~\ref{fig:feat_important_top10} illustrates the top 10 features by SHAP value, which measures how much each single feature pushed the model's prediction compared to the average prediction. The pausing behaviours (positions and durations), intensity measures (max, mean, and relative change), and harmonic-to-noise ratio (HNR) appear particularly important among prosodic features. For linguistic features, the grammatical structure linking to coreference used, some POS tags, and various word type ratios rank highly, which align well with systematic linguistic patterns, as demonstrated by \citet{Ngo2024}. The most important features include the number of long and medium pauses, the relative position of the longest pause, and the verb-followed-by-coref structure, all scoring near 1.0 on the importance scale, which aligned with the works in \cite{Elliott2018, Ngo2024} about pauses in repair initiation and its structure, respectively.

\begin{figure}[h]
  \centering
  \includegraphics[width=\columnwidth]{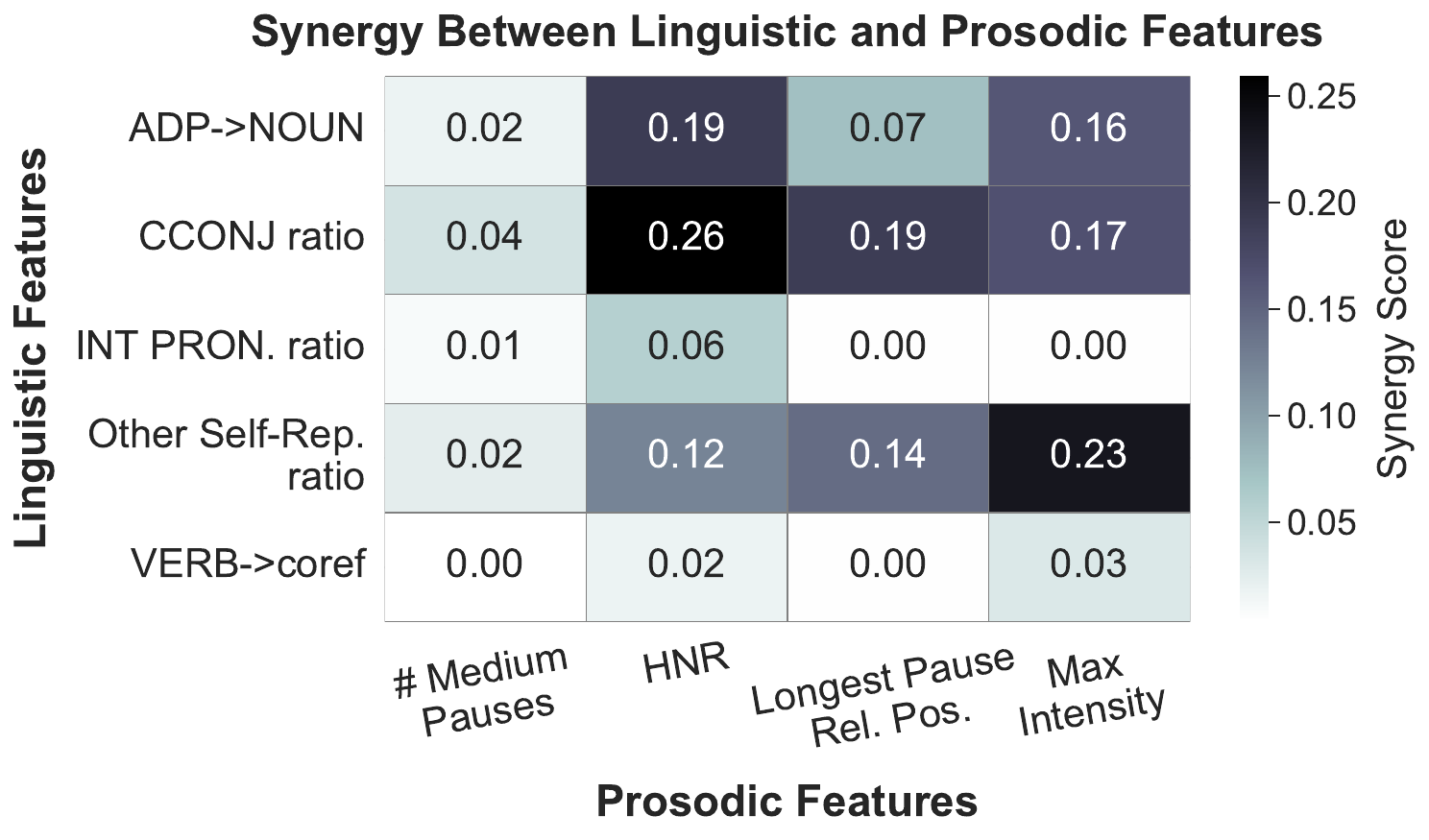}
  \caption{Handcrafted feature interaction analysis: Linguistic vs Prosodic}
  \label{fig:feat_interaction_heatmap}
\end{figure}

Figure~\ref{fig:feat_interaction_heatmap} displays the synergy \cite{Ittner2021} between linguistic and prosodic features, computed based on the SHAP interaction values. It reflects how complementary a pair of linguistic and prosodic features is in improving model performance, in which high synergy means that combining both features adds more value than what each of them contributes individually. These features do not always need to co-vary, but their combination brings useful information for the model. Coordinating conjunction ratio (CCONJ ratio) shows the strongest synergy (0.26) with harmonics-to-noise ratio (HNR), while other speaker self-repetition ratio has strong synergy (0.23) with maximum intensity. This suggests that certain grammatical patterns work closely with specific voice qualities, particularly how conjunctions interact with voice clarity and how self-repetition correlates with voice intensity. The results indicate that conversation involves a complex interplay between what we say (linguistic elements) and how we say it (prosodic elements), which is aligned with the Conversation Analysis work.

% Reflect the feature importance with the grounded literature review, important features + data samples + interpretation
% \paragraph{Cross-modal Feature Interaction.}

\paragraph{RQ4: Dialogue Micro Context Analysis.}
% For the past_context, we include: past + current, for future_context, we incude: current + future  , for current: only current, for Multi_ours, we include: past + current + future (as the current segment is the one we need to classify, it's always needed to have the current in input either solely or with past/future context)
To address RQ4, we experimented 4 scenarios of concatenating micro context, including: (1) \textbf{Past\textsubscript{Context}} - concatenated current input segment with the segments in the prior turns and cross-segment handcrafted features (past-related, Figure~\ref{fig:feat_design}); (2) \textbf{Future\textsubscript{Context}} - concatenated current input segment with the segments in the subsequent turns and handcrafted cross-segment features (future-related, Figure~\ref{fig:feat_design}); (3) \textbf{Current\textsubscript{Context}} - no context concatenation and used only current input segment features (Figure~\ref{fig:feat_design}); (4) \textbf{Multi\textsubscript{Ours}} - the full context scenario, where we concatenate current input segment with both the prior and subsequent segments and use full handcrafted feature set. For (1) and (4), we experimented with window\_length of 2 and max (the micro context are concatenated as much as possible until it reach maximum token limit) based on results from corpus analysis; for (3) only max was used, as repair solutions typically occur immediately within maximum 2 turns in this corpus.

\begin{table}[!h]
    \resizebox{\columnwidth}{!}{%
    \begin{tabular}{lcccc}
    \hline
    \textbf{Context} & \textbf{Win. len} & \textbf{Precision} & \textbf{Recall} & \textbf{F1-score} \\
    \hline
    % \multicolumn{5}{l}{\textbf{RQ1: Audio vs.\ Text Complementarity}} \\
    (1) Past\textsubscript{Context}         & 2        & $86.0\pm3.0$  & $78.4\pm5.4$ & $82.0\pm4.1$ \\
    (1) Past\textsubscript{Context}        & max       & $86.6\pm5.2$  & $81.0\pm6.1$ & $83.5\pm4.3$  \\
    % \rowcolor{gray!15}
    (2) Current\textsubscript{Context}        & -     & $84.6\pm3.8$  & $82.9\pm6.0$ & $83.6\pm4.4$  \\
    (3) Future\textsubscript{Context}        & max       & $84.00\pm1.53$  & $78.20\pm5.78$ & $80.18\pm2.52$  \\
    % \rowcolor{gray!15}
    \hline
    (4) Multi\textsubscript{Ours}   & 2     & $\mathbf{93.2\pm2.8}$ & $\mathbf{96.1\pm2.6}$ & $\mathbf{94.6\pm2.3}$  \\
    % \rowcolor{gray!30}
    (4) Multi\textsubscript{Ours}   & max  & $87.7\pm3.5$ & $89.1\pm5.3$ & $88.3\pm3.7$ \\
    \hline
    \end{tabular}%
    }
    % \captionsetup{justification=raggedright,singlelinecheck=false}
    \caption{Performance comparison across different micro context configurations}
    \label{tab:context_result}
\end{table} 

Table~\ref{tab:context_result} highlights the impact of different micro context configurations, in which incorporating surrounding segments from prior, and subsequent segments combining with the whole handcrafted feature set leads to the best overall performance, as also stated in Table~\ref{tab:overall_results}. Notably, our full context setting with smaller window\_length=2 achieves the highest results across all metrics, while concatenating to the maximum allowed token limits degrades the performance, with a drop of approximately 6.3 pp of F1-score, 9 pp of precision, and 4.1 pp of recall. It suggests that while surrounding context of input segment is helpful, overly long concatenation may introduce noise and irrelevant information to the model. In addition, integrating past or solely current segments yields moderate performance, with F1-scores ranging from approximately 80.2\% to 83.6\%, while future context integration results in the lowest scores, indicating that the upcoming dialogue can offer informative cues but less relevant than the prior and current input segments, which aligned with the nature of OIR sequence.
 
\section{Error Analysis}
\label{error_analysis}

To better interpret model performance, we analyze the False Negative (FN) instances, which are repair initiations that were misclassified as regular dialogue, to identify whether there are common patterns in these instances that our models struggle to predict, illustrated in Table~\ref{tab:fn_instances_analysis}. We compare these FN instances across our proposed multimodal model with the unimodal baselines by extracting representative dialogue samples for each model from test set and identifying their common linguistic and prosodic characteristics.

\begin{table*}[h!]
    \centering
    \renewcommand{\arraystretch}{1.0}
    \resizebox{0.9\textwidth}{!}{%
    \begin{tabular}{p{0.09\linewidth}|p{0.075\linewidth}|p{0.29\linewidth}|p{0.40\linewidth}|p{0.1\linewidth}}
        \hline
        \textbf{Model} & \textbf{\%Error} & \textbf{Samples} & \textbf{Patterns} & \textbf{OIR Type} \\
        \hline
        \multirow{3}{*}{\textbf{Text\textsubscript{Ling}}} 
            &  & \textit{(or a) triangle} & Vague, elliptical reference & RO \\
            & ~15\% & \textit{yes uh yes on the right side right? or ascending yes} & Disfluencies, vague interrogative & RO \\
            &  & \textit{yes the one with the protrusion} & Referential expression, lacks direct marker & RO \\
        \hline
        \multirow{5}{*}{\textbf{Audio\textsubscript{Pros}}} 
            &  & \textit{with a sunshade} & Short declarative, flat prosody & RO \\
            & ~24\% & \textit{uh but the platform sits that cuts the} & Flat intonation, short pauses in beginning & RO \\
            &  & \textit{Is it vertical?} & Question intonation, few short pauses & RO \\
            &  & \textit{ah and is his arm uh round but also a bit with angles?} & High pitch, question intonation, pauses mid-turn & RO \\
            &  & \textit{but what did you say at the beginning?} & Rising intonation, wide pitch range & RR \\
        \hline
        \multirow{3}{*}{\textbf{Multi\textsubscript{Ours}}} 
            &  & \textit{with a sunshade} & Short, declarative structure & RO \\
            & ~3.8\% & \textit{oh who so} & Declarative, high but flat pitch & RO \\
            &  & \textit{sorry again?} & Clear OIR but subtle prosodic signal & OR \\
        \hline
    \end{tabular}%
    }
    \caption{Samples of False Negative (FN) instances from unimodal and multimodal models with qualitative patterns. OR: open request; RR: restricted request; RO: restricted offer. The Dutch samples are translated to English by DeepL.}
    \label{tab:fn_instances_analysis}
\end{table*}

Our proposed model shows the lowest FN rate (~3.8\%) of the test set, compared to 15\% and 24\% on \textbf{Text\textsubscript{Ling}} and \textbf{Audio\textsubscript{Pros}}, respectively. \textbf{Text\textsubscript{Ling}} seems to struggle in detecting samples with vague references, especially in restricted offers, even when OIR syntactic forms like \emph{question mark} is present. Besides, \textbf{Audio\textsubscript{Pros}} tends to over-rely on pause structure and pitch contour even though important prosodic cues were presented. Short declaratives with flat intonation were often misclassified, suggesting the impact of missing syntactic form information in this model. Finally, our proposed multimodal failed with mostly short phrases and subtle prosodic signals, which are not strongly marked as an repair initiation. Considering the error across 3 types of repair initiations, it seems that only \textbf{Audio\textsubscript{Pros}} struggled with various types of repair initiations; the other 2 models misclassified on restricted offer and open request instances only. However, as this corpus is imbalanced between the 3 types of repair initiation, with a majority of restricted offers, it could be the potential reason.

\section{Conclusion \& Future Works}
\label{conclusion}
This work presents a novel approach for detecting repair initiation in Other-Initiated Repair (OIR) sequences within human-human conversation. It leverages automatically extracted linguistic and prosodic features grounded in Conversation Analysis theories. Our results demonstrate that incorporating handcrafted features significantly enhances detection performance compared to using only pretrained embedding models. Additionally, audio modality complements textual modality, improving detection performance across both pretrained embeddings and handcrafted features. Handcrafted feature analysis revealed both individual impact and complementary contributions between modalities. Key prosodic indicators include pause-related features, intensity, and harmonic-to-noise ratio (HNR), while important linguistic features involve grammatical patterns, POS tags, and lemma ratios. Synergy analysis demonstrates that features do not act independently; for example, coordinating conjunction usage shows strong synergy with HNR, and trouble source speaker self-repetition leads significantly to maximum intensity presence. These patterns highlight the nature of OIR sequences, in which how something is said modulates what is being said. 

Our results also highlight the importance of dialogue micro context in repair initiation detection: models using both prior and subsequent segments outperform those relying only on the target segment, reflecting the interactional structure crucial for OIR interpretation. However, overusing context can add noise and degrade performance. 

Finally, error analysis revealed that while the text-based model failed with vague references and disfluencies, the audio-based model was prone to misclassifying flat or subtle prosodic cues, which raised the need for a multimodal model. The proposed multimodal model mitigates these weaknesses, but it still struggles with short, minimally marked repair initiation that lacks both strong syntactic and prosodic cues. This work establishes foundations for conversational agents capable of detecting human repair initiation to avoid communication breakdowns.

Building on these insights, future work will explore the integration of visual features to more accurately model the embodied aspects of OIR sequences, as well as the development of multilingual and cross-context corpora to assess the robustness and generalizability of the detection approach.

\section*{Limitations}
\label{sec:limitations}

\paragraph{Dataset Limitations and Generalizability.} Due to the limited multimodal OIR-labeled corpora, our study utilized the only available multimodal OIR-labeled corpus, which is specific to Dutch language and referential object matching tasks. This specificity could limit the generalizability of our model across different OIR categories, languages, and conversation settings. Future works should test the model on more diverse datasets to validate its robustness and establish broader applicability.

\paragraph{Dataset Balancing and Class Distribution.} In natural conversation, repair initiation instances are much less frequent than regular dialogue. To enable robust model training and evaluation, we balanced repair initiation and regular dialogue samples across dyads. However, this balancing approach may affect the model’s performance in real-world settings where OIR sequences are rare, and therefore, the results should be interpreted with caution. Future work should evaluate the performance of models while maintaining the natural class distribution to assess practical applicability.

\paragraph{Adaptability in Real-time Processing.}
Despite the computational efficiency of our approach using handcrafted features compared to Large Language Models, several limitations remain for real-time adaptation. The feature extraction of some linguistic and prosodic features, such as coreference chains, requires additional computation with pretrained models, potentially introducing latency. Future work should explore real-time feature extraction pipelines and incremental processing architectures, while evaluating potential trade-offs between model complexity and real-time performance to make the system practical for CA systems.

\section*{Acknowledgments}
We thank the anonymous reviewers for their constructive feedback. Data were provided (in part) by the Radboud University, Nijmegen, The Netherlands. This work has been supported by the Paris Île-de-France Région in the framework of DIM AI4IDF. It was also partially funded by the ANR-23-CE23-0033-01 SINNet project. 

% This document has been adapted
% by Steven Bethard, Ryan Cotterell and Rui Yan
% from the instructions for earlier ACL and NAACL proceedings, including those for
% ACL 2019 by Douwe Kiela and Ivan Vuli\'{c},
% NAACL 2019 by Stephanie Lukin and Alla Roskovskaya,
% ACL 2018 by Shay Cohen, Kevin Gimpel, and Wei Lu,
% NAACL 2018 by Margaret Mitchell and Stephanie Lukin,
% Bib\TeX{} suggestions for (NA)ACL 2017/2018 from Jason Eisner,
% ACL 2017 by Dan Gildea and Min-Yen Kan,
% NAACL 2017 by Margaret Mitchell,
% ACL 2012 by Maggie Li and Michael White,
% ACL 2010 by Jing-Shin Chang and Philipp Koehn,
% ACL 2008 by Johanna D. Moore, Simone Teufel, James Allan, and Sadaoki Furui,
% ACL 2005 by Hwee Tou Ng and Kemal Oflazer,
% ACL 2002 by Eugene Charniak and Dekang Lin,
% and earlier ACL and EACL formats written by several people, including
% John Chen, Henry S. Thompson and Donald Walker.
% Additional elements were taken from the formatting instructions of the \emph{International Joint Conference on Artificial Intelligence} and the \emph{Conference on Computer Vision and Pattern Recognition}.

\bibliography{custom}

\appendix

\section{Dataset Details}
\label{sec:corpus_info}

\begin{figure}[h]
  \includegraphics[width=\columnwidth]{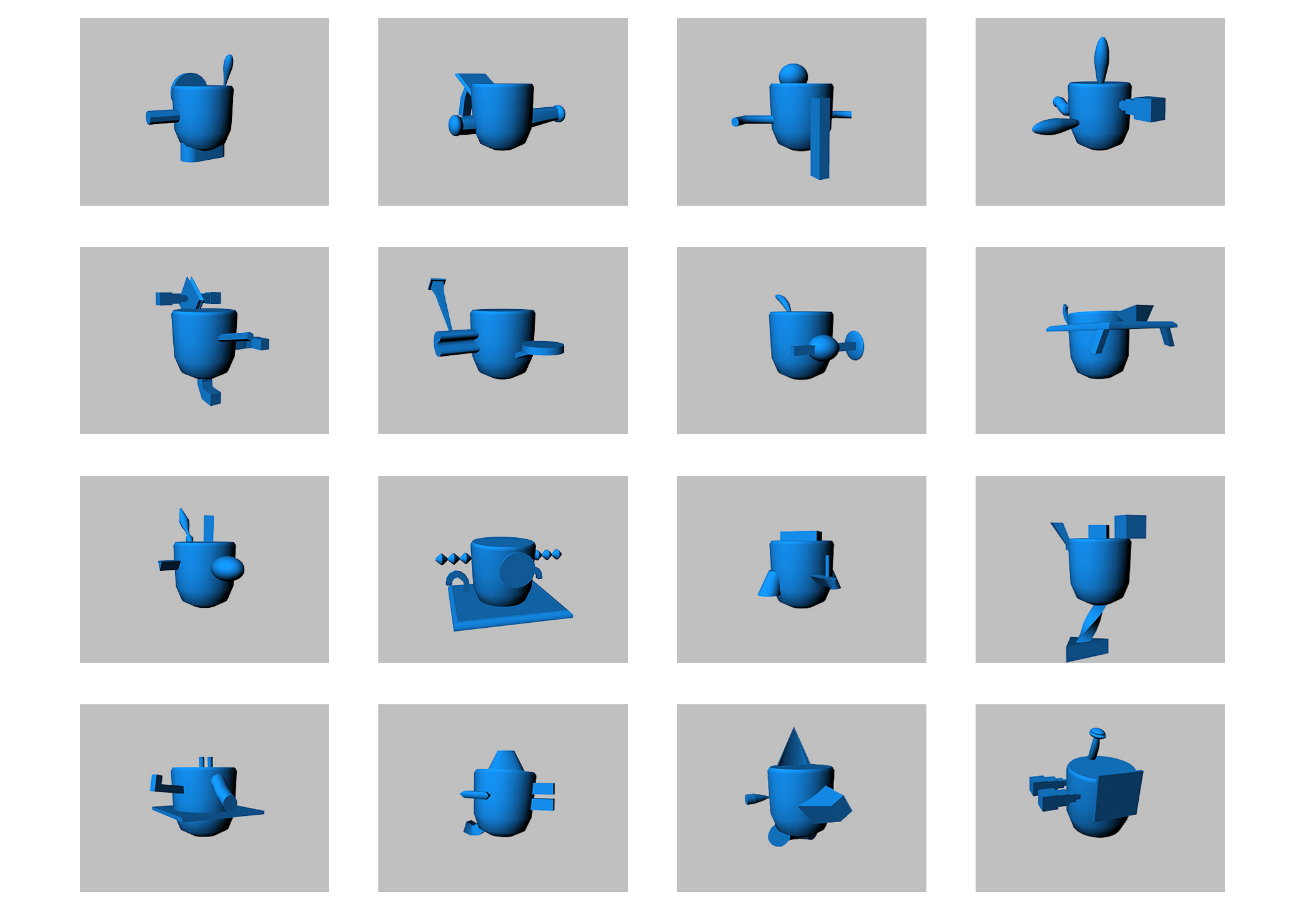}
  \caption{16 "Fribbles" were used in the object matching task \cite{Rasenberg2022, Eijk2022}.}
  \label{fig:fribbles}
\end{figure}

Figure~\ref{fig:fribbles} presents samples of 16 geometrical objects called "Fribbles" displayed on the participants' screens. Each dyad completed 6 rounds per session, resulting in 96 trials total. In each trial, participants alternated between Director and Matcher roles: the Director described a highlighted Fribble while the Matcher identified and confirmed the corresponding object by naming it loudly before proceeding to the next trial.

\label{sec: micro_context}

\section{OIR Types Examples}
\label{sec:oir_types_examples}
\begin{example}
    Open request sample    
    \begin{dialogue}
        \speak{\textbf{TS speaker}} op dat driehoek \hfill(TS) \\
        \textit{(on that triangle)}
        \speak{\textbf{Repair initiator}} wat zei je? \hfill(RI) \\
        \textit{(what did you say?)}
        \speak{\textbf{TS speaker}} op die driehoek \hfill(RS) \\
        \textit{(on that triangle)}
    \end{dialogue}
\end{example}

\begin{example}
    Restricted request sample 
    \begin{dialogue}
        \speak{\textbf{TS speaker}} deze heeft twee oren die aan de onderkant breder worden en een soort hanekam op zijn hoofd een kleintje \hfill(TS) \\
        \textit{(this one has two ears that widen at the bottom and a sort of cock's comb on its head a little one)}
        \speak{\textbf{Repair initiator}} maar wat zei wat zei je in het begin? \hfill(RI) \\
        \textit{(but what did you say at the beginning?)}
        \speak{\textbf{TS speaker}} een soort oren die aan de onderkant breder worden \hfill(RS) \\
        \textit{(a kind of ears that widen at the bottom)}
    \end{dialogue}
\end{example}

\begin{example}
    Restricted offer sample 
    \begin{dialogue}
        \speak{\textbf{TS speaker}} waarbij je dus op de bovenkant zo'n zo'n mini uh kegeltje hebt \hfill(TS) \\
        \textit{(where you have one of those mini uh cones on the top)}
        \speak{\textbf{Repair initiator}} oh ja die zo scheef naar achter staat? \hfill(RI) \\
        \textit{(oh yes which is so slanted backwards?)}
        \speak{\textbf{TS speaker}} ja precies \hfill(RS) \\
        \textit{(yes exactly)}
    \end{dialogue}
\end{example}

\section{Top 20 Important Features}
\label{sec:appendix_feature_important}
\begin{figure}[h]
  \includegraphics[width=\columnwidth]{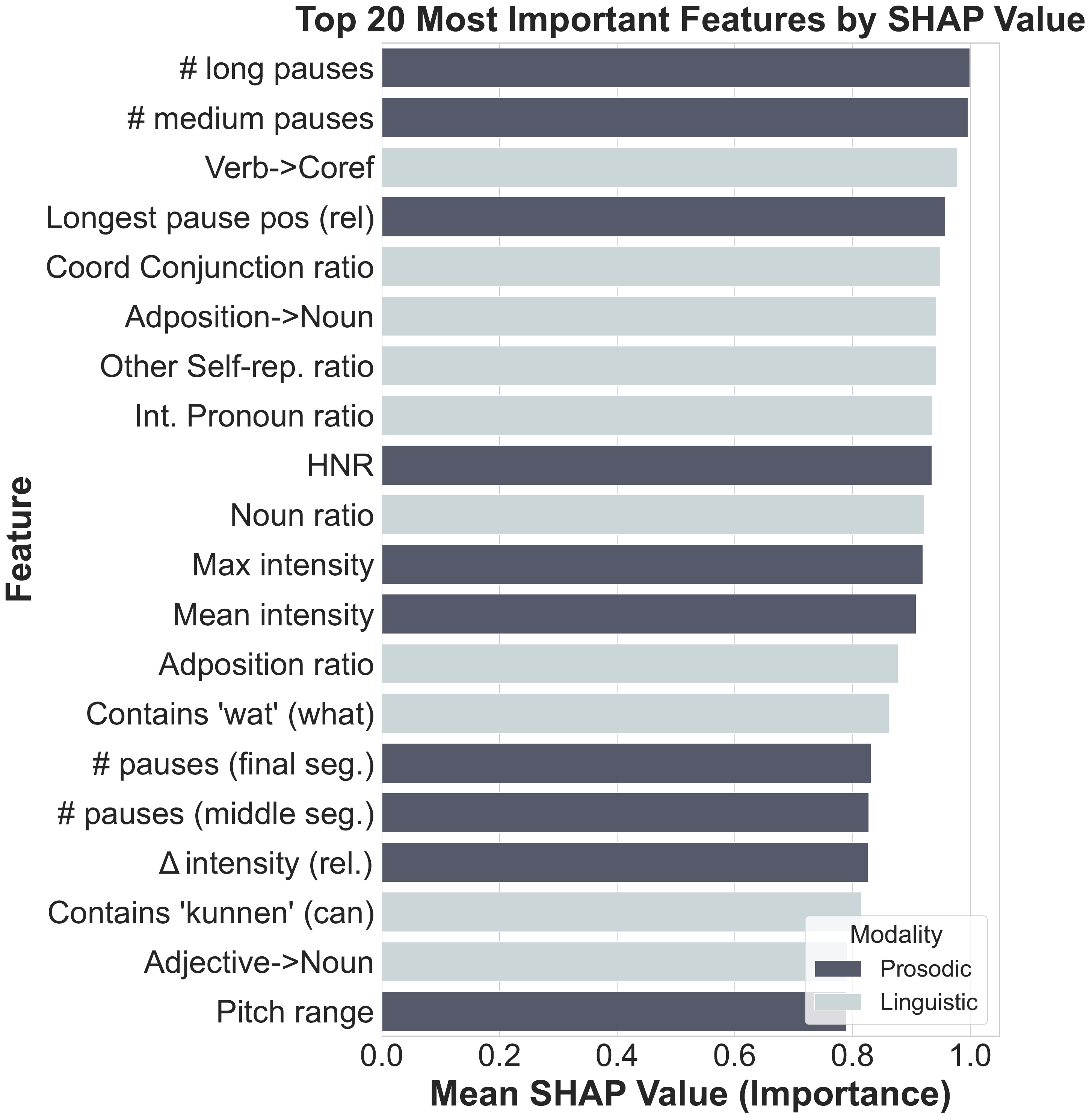}
  \caption{Top 20 most contributed features by SHAP values.}
  \label{fig:top_20_important_feat}
\end{figure}
\newpage

\section{Dialogue Micro Context}
\label{sec: micro_context}
\begin{figure}[h]
  \includegraphics[width=\columnwidth]{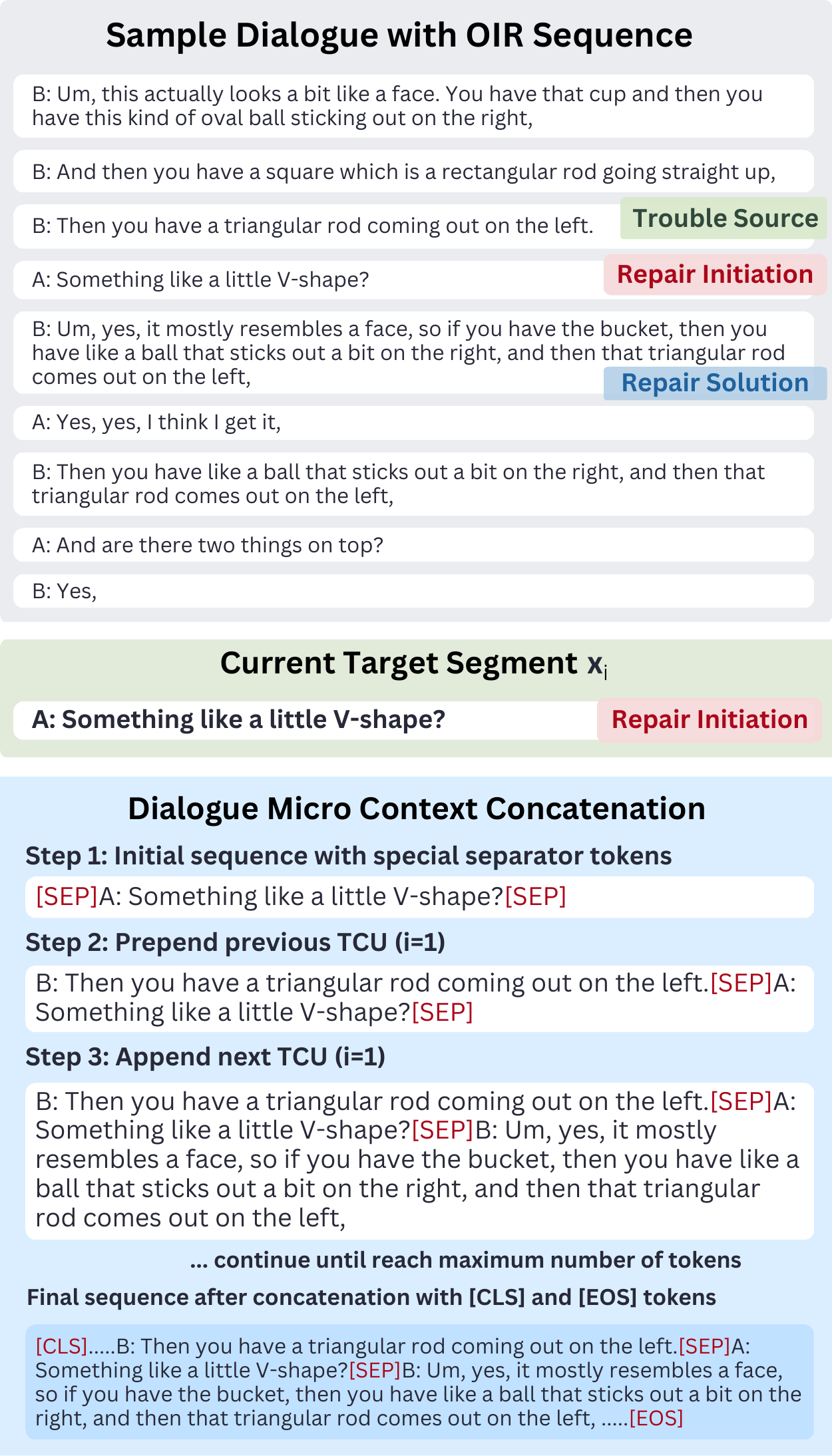}
  \caption{Dialogue micro context concatenation approach. \textit{Micro context} refers to the immediate conversational environment, including the prior and the subsequent segments of the current target segment in dialogue \cite{Schegloff1987}.}
  \label{fig:dialogue_context_concatenation}
\end{figure}

\FloatBarrier
\section{Detailed Linguistic Features}
\label{sec:appendix_linguistic}

Table~\ref{tab:linguistic_features} summarizes the handcrafted feature set that were automatically extracted using the approach proposed in \citet{Ngo2024}'s work. 

\begin{table*}[!ht]
\small
\renewcommand{\arraystretch}{1.3}
\resizebox{\textwidth}{!}{%
    \begin{tabular}{|p{3.2cm}|p{2.5cm}|p{4.5cm}|p{5.2cm}|}
    \hline
    \textbf{Level} & \textbf{Feature Group} & \textbf{Feature Type(s)} & \textbf{Description} \\
    \hline
    \multirow{2}{*}{Segment-level} 
    & POS tags sequence & \texttt{POS tag bigrams}, \texttt{POS tag ratios} & Binary features for frequent POS tag bigrams (e.g., \texttt{PRON\_Prs$\rightarrow$VERB}, \texttt{VERB$\rightarrow$COREF}); POS tags frequency ratios computed per segment. \\
    & Lemma & \texttt{contains\_lemma} (e.g., \texttt{nog}, \texttt{kunnen}) & Binary indicators for presence of high-frequency lemmas relevant to different type of repair initiation. \\
    & Question form & \texttt{ends\_with\_question\_mark} & Binary feature indicating whether the segment ends with a question mark. \\
    & Non-verbal action & \texttt{contains\_laugh}, \texttt{contains\_sigh}, etc. & Binary features for transcribed non-verbal actions like \#laugh\#, \#sigh\#, etc. \\
    \hline
    \multirow{2}{*}{\shortstack{Cross-segment level\\(prior turns related)}}
    & Repetition from previous turn & \texttt{other\_repetition\_ratio} & Ratio of tokens in the current segment that are repeated from the other speaker’s previous turn relative to total segment length. \\
    & Coreference from previous turn & \texttt{coref\_used\_ratio} & Ratio of coreference phrases (e.g., pronouns or noun phrases referring to previous turn) relative to total segment length. \\
    \hline
    \multirow{2}{*}{\shortstack{Cross-segment level\\(subsequent turns related)} }
    & Repair solution TSS self-repetition & \texttt{other\_speaker\_self\_rep\_ratio} & Ratio of self-repetition in the turn following the repair initiation. \\
    & Repair solution TSS other-repetition & \texttt{other\_speaker\_other\_rep\_ratio} & Ratio of other-repetition in the turn following the repair initiation \\
    \hline
    \end{tabular}
}
\caption{Summary of linguistic feature set used for modeling repair initiation. The full POS tag list includes: ADJ (adjectives), ADP (prepositions and postpositions), ADV (adverbs), AUX (auxiliaries, including perfect tense auxiliaries "hebben" (\textit{to have}), "zijn" (\textit{to be}); passive tense auxiliaries "worden" (\textit{to become}), "zijn" (\textit{to be}), "krijgen" (\textit{to get}); and modal verbs "kunnen" (\textit{to be able, can}), "zullen" (\textit{shall}), "moeten" (\textit{must}), "mogen" (\textit{to be allow})), CCONJ (coordinating conjunctions such as "en" (\textit{and}), "of" (\textit{or})), DET (determiners), INTJ (interjections), NOUN (nouns), PRON\_Dem (demonstrative pronouns), PRON\_Int (interrogative pronouns), PRON\_Prs (personal pronouns), PUNCT (punctuation), SYM (symbols), and VERB (verbs). The considered common lemma includes: \textit{wat} (what), \textit{kunnen} (can), \textit{zitten} (to sit/set), \textit{zijn} (to be), \textit{nog} (yet/still), \textit{wachten} (to wait), \textit{aan} (on/to/at/in/by/beside/upon). And the transcribed non-verbal actions includes: \textit{laughs}, \textit{sighs}, \textit{breath}, and \textit{mouth noise}.}
\label{tab:linguistic_features}
\end{table*}
\FloatBarrier   

\section{Detailed Prosodic Features}
\label{sec:appendix_prosodic}

\begin{table}[h]
\centering
\small
\renewcommand{\arraystretch}{1.1}
\resizebox{\textwidth}{!}{%
    \begin{tabular}{|c|l|p{3.5cm}|p{6.5cm}|}
    \hline
    \textbf{Level} & \textbf{Feature Group} & \textbf{Feature Type} & \textbf{Description} \\
    \hline
    \multirow{6}{*}{Segment-level} 
    & Pitch features 
    & \texttt{min}, \texttt{max}, \texttt{mean}, \texttt{std}, \texttt{range}, \texttt{num\_peaks} 
    & Extracted from voiced frames; outliers removed; peaks from smoothed contour \\
    & Pitch dynamics 
    & \texttt{slope}
    & Captures pitch variation within segment. \\
    & Intensity features 
    & \texttt{min}, \texttt{max}, \texttt{mean}, \texttt{std}, \texttt{range} 
    & Computed from nonzero intensity frames; reflects loudness. \\
    & Voice quality 
    & \texttt{jitter}, \texttt{shimmer}, \texttt{hnr} 
    & Reflects vocal fold irregularity and breathiness. \\
    & Pause features 
    & \texttt{num}, \texttt{durations}, \texttt{short/med/long}, \texttt{positional counts}, \texttt{rel\_longest} 
    & Pause detection using adaptive thresholds; categorized by duration and position. \\
    & Speech timing 
    & \texttt{rate}, \texttt{articulation\_rate}, \texttt{duration} 
    & Segment length and estimated speech rate (e.g., syllables/sec). \\
    \hline
    \multirow{3}{*}{\shortstack{Cross-segment level\\(both prior and\\subsequent related)} }
    & Transition features 
    & \texttt{end\_slope}, \texttt{start\_slope}, \texttt{transition} 
    & Pitch slope difference across segment boundaries (prev$\rightarrow$cur, cur$\rightarrow$next); in semitones/sec. \\
    & Baseline comparison 
    & \texttt{z\_score}, \texttt{rel\_change}, \texttt{range\_pos} 
    & Comparison to speaker's pitch/intensity baseline. \\
    & Latency 
    & \texttt{TS$\rightarrow$RI}, \texttt{RI$\rightarrow$RS} 
    & Silence duration between trouble source and repair initiation, repair initiation and repair solution. \\
    \hline
    \end{tabular}%
}
% \captionsetup{width=5\linewidth, justification=centering}
% \caption{Summary of prosodic feature set used for modeling repair initiation.}
\caption{\makebox[0.6\textwidth][c]{Summary of prosodic feature set used for modeling repair initiation.}}
\label{tab:prosody_features}
\end{table}

\end{document}